\documentclass{article}

\usepackage{arxiv}
\usepackage[utf8]{inputenc} 
\usepackage[T1]{fontenc}    
\usepackage{hyperref}       
\usepackage{url}            
\usepackage{booktabs}       
\usepackage{amsfonts}       
\usepackage{amsmath}        
\usepackage{nicefrac}       
\usepackage{microtype}      
\usepackage{graphicx}
\usepackage{float}
\graphicspath{{./images/}}

\title{Medical Image Classification on Imbalanced Data Using ProGAN and SMA-Optimized ResNet: Application to COVID-19}

\author{
  Sina Jahromi \\
  Department of Mechatronics, School of Intelligent Systems\\
  College of Interdisciplinary Science and Technology\\
  University of Tehran\\
  Tehran, Iran\\
  \texttt{sina.jahromi@ut.ac.ir} \\
  \And
  Farshid Hajati\\
  School of Science and Technology\\
  Faculty of Science, Agriculture, Business and Law\\
  University of New England\\
  Armidale, NSW 2350, Australia\\
  \texttt{fhajati@une.edu.au} \\
  \And
  Alireza Rezaee \\
  Department of Mechatronics, School of Intelligent Systems\\
  College of Interdisciplinary Science and Technology\\
  University of Tehran\\
  Tehran, Iran\\
  \texttt{arrezaee@ut.ac.ir} \\
  \And
  Javaher Nourian \\
  Department of Mechatronics, School of Intelligent Systems\\
  College of Interdisciplinary Science and Technology\\
  University of Tehran\\
  Tehran, Iran\\
  \texttt{javahernourian@ut.ac.ir}
}

\begin{document}
\maketitle

\begin{abstract}
The challenge of imbalanced data is prominent in medical image classification. This challenge arises when there is a significant disparity in the number of images belonging to a particular class, such as the presence or absence of a specific disease, as compared to the number of images belonging to other classes. This issue is especially notable during pandemics, which may result in an even more significant imbalance in the dataset. Researchers have employed various approaches in recent years to detect COVID-19 infected individuals accurately and quickly, with artificial intelligence and machine learning algorithms at the forefront. However, the lack of sufficient and balanced data remains a significant obstacle to these methods. This study addresses the challenge by proposing a progressive generative adversarial network to generate synthetic data to supplement the real ones. The proposed method suggests a weighted approach to combine synthetic data with real ones before inputting it into a deep network classifier. A multi-objective meta-heuristic population-based optimization algorithm is employed to optimize the hyper-parameters of the classifier. The proposed model exhibits superior cross-validated metrics compared to existing methods when applied to a large and imbalanced chest X-ray image dataset of COVID-19. The proposed model achieves 95.5\% and 98.5\% accuracy for 4-class and 2-class imbalanced classification problems, respectively. The successful experimental outcomes demonstrate the effectiveness of the proposed model in classifying medical images using imbalanced data during pandemics.
\end{abstract}


\section{INTRODUCTION}

In this research, an attempt has been made to develop a method based on modern artificial intelligence that can detect people infected with this virus as quickly as possible and at the lowest cost. One of the benefits of early detection of people infected with this virus is to prevent its further spread. Also, another of its other benefits is the possibility of saving the lives of infected people before their condition worsens. The current research uses an ImageNet pre-trained ResNet50V2 model to classify CXR images into four labels (Normal, Viral Pneumonia, COVID-19, Lung Opacity). One of the main challenges in using this method to classify images is the lack of sufficient data to train the neural network. Furthermore, the inequality of classes’ frequencies causes an unfavorable bias in the classifier network during the training/application stages.

Generative adversarial networks (GANs) have been used to solve the mentioned challenges. In general, adversarial generative networks are trained to imitate the distribution of the input data of the network in the output (ideally, without the occurrence of mode collapse, which leads to low diversity in the distribution of the output data). With the help of this type of neural network, as well as the help of limited existing images, new synthetic CXR images have been produced. It is demonstrated that using these synthetic images improves the classifier’s performance. Training GANs is usually a formidable task due to the intrinsic instability of such networks and being highly dependent on networks’ structures. Thus, the produced images would have a low quality (which means the GAN imitates fewer features of CXR images in its output) \cite{Arjovsky2017WGAN,Mahdizadehaghdam2019SparseGAN,Gulrajani2017ImprovedWGAN}. In order to deal with this challenge, a progressive training method has been used: this type of GAN is called ProGAN, which not only improves the quality of produced images, but also helps to avoid network bias \cite{Karras2018ProGAN}. Additionally, to improve the classification, Slime Mould Algorithm (SMA) has been used to optimize the hyper-parameters of the classifier network. The basis of the function of this optimizer, which is a population-based meta-heuristic algorithm, is inspired by the oscillation state of slime mould in nature. Eventually, the results indicate the classifier network’s performance improvement in different classification modes. In essence, The core contribution of this paper is the proposal of a novel method to tackle the problem of limited data availability in poorly populated classes. The method involves using a customized ProGAN for each class as an image augmentation technique and an SMA optimization algorithm to improve the classifier network's performance by optimizing the hyperparameters of the network. To counter the issue of class imbalance in the dataset, a weighted approach is suggested that combines the synthesized images from the ProGAN with the real CXR images. By doing so, it is ensured that the frequency of each class is represented more evenly in the training dataset. As of writing this paper, the proposed methods were not used in other studies.

\section{RELATED WORKS}

Roy et al.\ \cite{Roy2022SVDCLAHE} developed a framework to address the challenge of imbalanced data in COVID-19 detection using chest X-ray (CXR) images. Their approach included two main components: an image contrast enhancement technique called Singular Value Decomposition-Contrast Limited Adaptive Histogram Equalization (SVD-CLAHE), and a balanced loss function called Balanced Weighted Categorical Cross-Entropy (BWCCE). The SVD-CLAHE technique was used to improve the contrast of CXR images with poor contrast. This enabled the classifier to better learn the features of these images. The BWCCE loss function was used to improve the classifier's performance in dealing with highly imbalanced dataset, which are common in conventional COVID-19 datasets. Overall, the proposed framework achieved impressive results with 94\% accuracy, 95\% F1-score, 96\% precision, and 95\% recall. Calderon-Ramirez et al.\ \cite{Calderon2021MixMatch} endeavored to address the obstacles related to imbalanced COVID-19 CXR datasets that contain a very limited number of samples. They assessed the effectiveness of MixMatch, a semi-supervised deep learning architecture, in this regard. Furthermore, they adopted a straightforward technique of assigning greater weight to the observations linked with under-represented classes during loss computation. Ultimately, they demonstrated that the employed algorithm augmented classification accuracy by 18\% compared to the non-balanced MixMatch algorithm. Nour et al.\ \cite{Nour2020DeepFeatures} introduced  a novel medical diagnosis for Covid-19 infection detection based on deep feature and Bayesian optimization. Their experiments were conducted on a publicly available COVID-19 dataset containing 2,905 images: 219 COVID images, 1,341 normal images, and 1,345 viral pneumonia images. They used an entirely customized CNN consisting of five convolutional layers and a deep feature extractor, which was trained from scratch; no transfer learning methods were used for training the CNN. Then, the extracted deep features were fed into three machine learning algorithms: K-nearest neighbor, support vector machine (SVM), and decision tree. Moreover, the hyperparameters of the neural network were optimized using Bayesian optimization. Eventually, they achieved 98.97\% as the best-acquired accuracy with the SVM classifier amongst conducted experiments \cite{Nour2020DeepFeatures}. 

As a comparative study among conventional feature extractor models, an study aiming to detect diabetic retinopathy was conducted by Wan et al.\ throughout evaluating several convolutional neural networks (AlexNet, VGGNet, GoogleNet, and ResNet) and utilizing a greater CXR dataset. Transfer-learning and hyper-parameter tuning methods were utilized in the training phase of the models to address three different challenges: classification, segmentation, and detection. An open-source Kaggle dataset called Diabetic Retinopathy Detection, which contained 25,810 normal, 2,443 mild, 5,292 moderate, 873 severe, and 708 proliferative images was used. Finally, the VGGNet achieved the highest accuracy with a value of 95.86\%. Consequently, the AlexNet achieved the lowest accuracy with a value of 89.75\%. It should be noted that the mentioned accuracy was acquired after the hyperparameter-tuning \cite{Wan2018DR}. Zinair et al.\ \cite{Zunair2021CycleGAN} proposed a novel method to tackle the challenges of imbalanced and scarce datasets during a new disease outbreak. They utilized deep learning to generate synthetic data that would enhance the performance of classifiers while also preserving the anonymity of medical data. During the adversarial phase of training, the authors used a CycleGAN model, which learned unpaired image-to-image translation. They discovered that oversampling poorly populated classes (in their case, the positive COVID-19 class) with synthetic data could substantially improve the classifier's ability to detect positive cases by reducing dataset skewness. Overall, their integrated deep learning framework is a promising solution for enhancing the accuracy of classifiers when working with imbalanced and scarce medical datasets. Zhao et al.\ \cite{Zhao2021EHSSA} presented a creative novel approach called EHSSA which thoroughly enhanced the performance of the Salp Swarm Algorithm. The enhanced algorithm is then used to tackle the local optimum problem using multi-threshold image segmentation (MIS) techniques. Moreover, their algorithm was also used to detect breast cancer using microscopic tissue images, leading to outstanding results. Rahman et al.\ \cite{Rahman2021Enhancement} compiled a large dataset containing 18479 CXR images in total. The compiled dataset consists of 3 labels with a frequency of 8851, 6012, and 3616 for Normal, non-Covid lung infection, and COVID-19(+), respectively. Afterward, a novel U-Net model was trained for lung section segmentation (The results of the novel U-Net is also compared with the conventional one). Finally, six deep CNNs were used to classify the segmented images (the utilized networks are: ResNet18, ResNet50, ResNet101, InceptionV3, DenseNet201, and ChexNet). Their results show that the detection accuracy using the not-segmented lung images is slightly better than the segmented ones. However, the network’s reliability is significantly higher when segmented lung images are used (this fact is concluded using the Score-CAM technique). As another research which a comparatively large dataset is utilized, A. Haghanifar et al.\ \cite{Haghanifar2022COVIDCXNet} used the ChexNet model to create a new classifier called COVID-CXNET to extract relevant and adequate features from CXR images efficiently. The main reason stated by the authors for choosing this approach is that the simple models and conventional transfer learning (mainly, the exploitation of pre-trained models) approaches can merely provide irrelevant features for the required application. Therefore, ChexNet model is used to create a robust classifier network. They have also created a large dataset of CXR images by concatenating several datasets (consisting of 10380 CXR images). Finally, they achieved 98.68\% accuracy for evaluating the model using 683 sample images assorted into Normal and Covid labels. To address the issue of imbalanced datasets and limited data, Badawi et al.\ \cite{Badawi2021Balanced} created a unified and balanced dataset that included 15,000 chest X-ray images from 11 publicly available datasets \cite{Wang2019ChestXray,Kermany2018Cell,Winther2020ImageRepo,Vaya2020BIMCV,Cohen2020Dataset,Haghanifar2022COVIDCXNet,Shams2020Mendeley,Chung2020Figure1,Chung2020ActualMed,ItalianSocietyRadiology2021,TwitterChestImaging}. They also used transfer learning techniques to train different classifiers with this balanced dataset. Their approach resulted in an impressive 99.62\% accuracy for binary classification and 95.48\% for multi-class classification (Pneumonia, Normal, and COVID) using a pretrained VGG16 network.

Unfortunately, their study was published after our research was completed. However, for future work on developing our method described in this paper, we believe that combining our approach with their compiled dataset would undoubtedly enhance the performance of classifiers significantly.

\section{MATERIAL AND METHODS}

\subsection*{3.1 DATA}

This study uses a publicly available dataset called ``COVID-19 Radiography Database'' for training and testing of the proposed model \cite{Rahman2021Enhancement,Chowdhury2020AI}. The dataset’s providers collected the images from nine different sources. The dataset contains 21,165 .png images with various resolutions. This largest available COVID-19 dataset is assorted into four labels: COVID-19, Normal, Viral Pneumonia, and Lung Opacity. The frequency and ratio, the label’s frequency divided by total number of samples, of each label are listed in Table~\ref{tab:label_freq}.

\begin{table}[H]
\centering
\caption{Frequency and ratio of each label available in the utilized dataset}
\label{tab:label_freq}
\begin{tabular}{lcc}
\toprule
Labels & Frequency & Ratio \\
\midrule
COVID-19 & 3,616 & 0.1708 \\
Normal & 10,192 & 0.4815 \\
Viral Pneumonia & 1,345 & 0.0635 \\
Lung Opacity* & 6,012 & 0.2841 \\
\midrule
Total & 21,165 & 1.00 \\
\bottomrule
\end{tabular}

\vspace{0.5em}
\small * The Lung Opacity label contains images of non-covid lung infections.
\end{table}

According to above table, there is a severe imbalance between labels’ frequency, directly affecting the classifier network (i.e., the labels with higher frequency can bias the classifier network and make it learn their features dominantly). A weighted synthetic image injection approach is used to deal with this challenge. 

In what follows, this approach is elaborated on:

\begin{equation}
n_{\text{label, with maximum freq.}} = n_{\text{Normal}}
\label{eq:2}
\end{equation}

\begin{equation}
n_{\text{label, complementary freq. (cf)}} = n_{\text{Normal}} - n_{\text{label}}
\label{eq:3}
\end{equation}

\begin{equation}
n_{\text{total, cf}} = \sum_{\text{labels}} n_{\text{label, cf}}
\label{eq:4}
\end{equation}

\begin{equation}
W_{\text{label, injection weight (iw)}} = \frac{n_{\text{label, cf}}}{n_{\text{total, cf}}} \times a_{\text{label}} \quad , \text{where label} = [\text{Covid-19, Viral Pneumonia, Lung Opacity}]
\label{eq:5}
\end{equation}

In equation (\ref{eq:5}), the $a_{\text{label}}$ is a fine-tuning parameter that can be defined label-wise. In the current research, the value of this parameter is considered one in all experiments. Thereby, the complementary frequency of each label and its respective injection weight are shown in Table~\ref{tab:comp_freq} (computed using equations (\ref{eq:2}) to (\ref{eq:5})):

\begin{table}[H]
\centering
\caption{Complementary Frequency \& Injection Weight of each label}
\label{tab:comp_freq}
\begin{tabular}{lcc}
\toprule
Labels & Complementary Frequency ($n_{\text{label, cf}}$) & Labels Injection Weight ($W_{\text{label, iw}}$) \\
\midrule
Covid-19 & 6576 & 0.34 \\
Viral Pneumonia & 8847 & 0.45 \\
Lung Opacity & 4180 & 0.21 \\
\bottomrule
\end{tabular}
\end{table}

In the optimization step of the research, an overall synthetic images injection ratio (SIIR), which is calculated by equation (\ref{eq:6}), is defined:

\begin{equation}
\text{SIIR} = \frac{N_{\text{fake (f)}}}{N_{\text{fake (f)}} + N_{\text{real (r)}}}
\label{eq:6}
\end{equation}

According to equation (\ref{eq:6}), the desired total number of synthetic images ($N_f$) is calculated by using equation (\ref{eq:7}):

\begin{equation}
N_f = \frac{\text{SIIR} \times N_r}{1 - \text{SIIR}}
\label{eq:7}
\end{equation}

Therefore, the desired number of synthetic images to be injected into the real dataset for each label ($n_{\text{label, fake (f)}}$) is calculated as follows (equation (\ref{eq:8})):

\begin{equation}
n_{\text{label, f}} = N_f \times W_{\text{label, iw}} \quad , \text{where label} = [\text{Covid-19, Viral Pneumonia, Lung Opacity}]
\label{eq:8}
\end{equation}

By the time $n_{\text{label, f}}$ is calculated through the above equation, it is ensured that injecting synthetic images is done in a way that there is an inverse relationship between number of injected synthetic images and frequency of real labels. In other words, more synthetic images are added to the real labels with less frequency. Consequently, the imbalance challenge is placated. The total number of injected synthetic images will be constrained by a single parameter which then is used to optimize the ratio of synthetic images available in the combined dataset.

Throughout the training procedure, the generated synthetic images are only added to the train and validation splits. Consequently, the computed metrics for test split merely correspond to the classification of real images.

To classify medical images, we employed adversarial networks in our research. These networks comprise two blocks: the Discriminator and the Generator, both built using a multi-layer perceptron structure. The training process involves adjusting the weights of the generator block so that the distribution of the generated synthetic data aligns with that of the input data. Initially, the generator randomly selects points from a latent space as input, and then aims to establish a connection between the distribution of those points and the distribution of the generated synthetic data based on the distribution of input data. The discriminator block calculates the probability of whether the data is real data or the generator (synthetic). 

\section{METHODOLOGY}

The chosen approach for performing experiment 1 in this study consists of 3 steps:
\begin{itemize}
    \item Training ProGAN for each class to generate a synthetic dataset of CXR images.
    \item Optimizing Learning Rate, SIIR ratio, and the dropout rate for an ImageNet pre-trained ResNet50V2.
    \item Training the final classifier network using the optimized values.
\end{itemize}

\begin{figure}[H]
\centering
\includegraphics[width=0.8\linewidth]{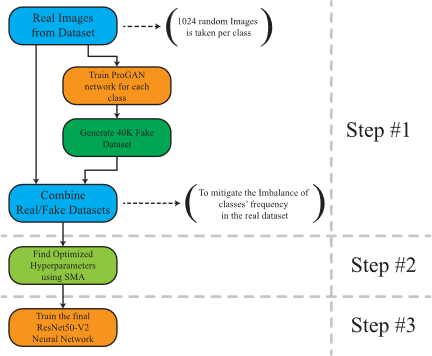}
\caption{Basic diagram of the performed methodology.}
\label{fig:2}
\end{figure}

In another method used for experiment 2, a VGG16 network is used to classify images. This method consists of injecting synthetic images into real dataset with different ratios (from 0.00 to 0.90 with a step of 0.1). Hence, the changes in classifier metrics could be easily obtained as different injection ratios.

To ensure the validity of the results in classification tasks \cite{WongYeh2020Kfold}, in both experiments, 10-fold cross-validation is used to compute the final metrics of the experiments’ classifier. The reported metrics are the average of the calculated metrics of each fold using a Testing Set. Subsequently, the remaining images (made by concatenating blue squares in each fold) will be divided into Training Set and Validation Set in each fold (the ratio of train and validation splits is a constant with a value of 0.15).

\subsection*{SMA Hyperparameters}

According to several studies, SMA–a metaheuristic population-based multi-objective optimization algorithm–showed significant performance for optimizing varied types of problems: mechanical and mathematical are cases in point. Therefore, after several experiments, the empirical values represented in Table~\ref{tab:sma_params} are used as the training parameters for the utilized SMA. Albeit, the selected values are not the most efficient ones; thus, they could be improved to enhance the performance of the optimization algorithm.

\begin{table}[H]
\centering
\caption{SMA parameters}
\label{tab:sma_params}
\begin{tabular}{ll}
\toprule
Values/Descriptions & Parameters \\
\midrule
15 & Population Size \\
250 & Epochs \\
15 & Problem Epochs (*) \\
Loss of the classifier network & Objective Function \\
min & Optimization Method (min/max) \\
{[}Learning Rate, Dropout Rate, FIIR{]} & Targets \\
{[}1e-3, 0.25, 0.5{]} & Upper bounds of targets \\
{[}1e-5, 0.05, 0{]} & Lower bounds of targets \\
128 & Batch Size (*) \\
\bottomrule
\end{tabular}

\vspace{0.5em}
\small (*) The parameters in these rows are not directly related to SMA (these values are related to the classifier network and hardcoded in the objective function as a constant value).
\end{table}

\subsection*{ProGAN Structure \& Parameters}

The original ProGAN network generates megapixel images with a size of 1024×1024 \cite{Karras2018ProGAN}. In this research, the same network is customized to generate 224×224 images. It should be noted that the indicated structure is for the final stage of progressive training. In the first stage, the generator outputs 7×7 images, and the critic feeds with images of the same size. Throughout each progression stage, the size of the images is doubled. Furthermore, the batch size and epoch count are different in each stage. The hyper-parameters used for the customized ProGAN network are illustrated in Table~\ref{tab:progan_params}:

\begin{table}[H]
\centering
\caption{The customized ProGAN hyper-parameters}
\label{tab:progan_params}
\begin{tabular}{ll}
\toprule
Values/Descriptions & Parameters \\
\midrule
6 & Progression Iteration Count \\
{[}256, 128, 32, 16, 16, 8{]} & Batch size of each progression stage (from 1 to 6) \\
3 & Image Channels Count \\
1e-3 & Learning Rate \\
112×1 & Latent Space Size \\
{[}250, 300, 350, 400, 450, 500{]} & Epochs Count in each progression stage \\
4096 & Real Images Count \\
5 & Critic’s iteration counts per generator’s \\
WGAN-GP \cite{Gulrajani2017ImprovedWGAN} & Loss’s Calculation Mathematical Model \\
\bottomrule
\end{tabular}
\end{table}

The main reason for varying epochs count in each progression stage is that the higher the resolution of the images, the higher the complexity and number of features embedded in them; Therefore, when generating images with a larger size, more epochs are needed to ensure that the ProGAN network learns them adequately. On the other hand, the main reason for varying batch sizes in each progression stage is the amount of required RAM. The higher the batch size is, the more RAM capacity is required. Moreover, in the final stages of progression, due to the generation of larger images, very high the amount of RAM would be required. Consequently, the batch size should be reduced when a high-capacity RAM is unavailable.

\begin{figure}[H]
\centering
\includegraphics[width=0.8\linewidth]{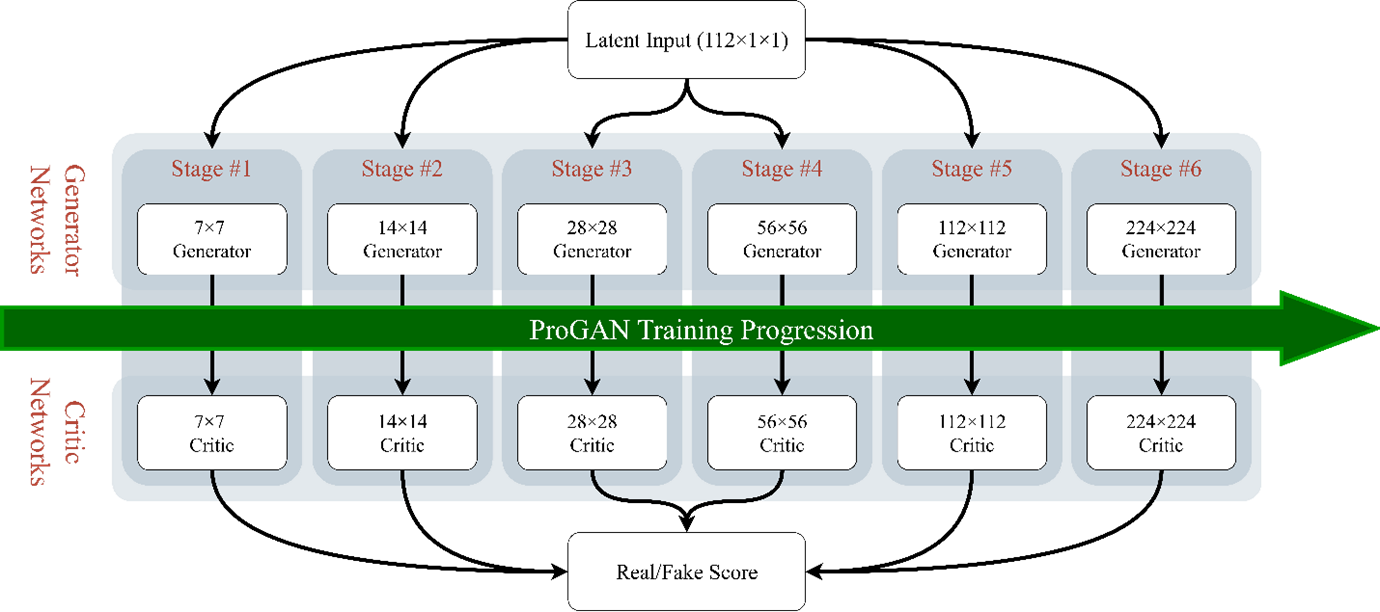}
\caption{General structure of generator output size and critic input size networks throughout training procedure.}
\label{fig:3}
\end{figure}

In Figure~\ref{fig:3}, a simple graphical representation of the training procedure is represented. It should be noted that the generator and critic networks of each stage are trained separately.

The structure of the customized generator network is shown in Table~\ref{tab:gen}:

\begin{table}[H]
\centering
\caption{Structure of the generator network of utilized ProGAN at stage \#6}
\label{tab:gen}
\begin{tabular}{lcccccc}
\toprule
Layer Output Dimensions & Layer Input Dimensions & Activation Functions & S & P & K & Layer Type \\
\midrule
112×1×1 & 112×1×1 & - & - & - & - & Latent Input \\
224×7×7 & 112×1×1 & LeakyReLu(0.2) & 1 & 0 & 7 & TConv2D \\
224×7×7 & 224×7×7 & LeakyReLu(0.2) & 1 & 1 & 3 & Conv2D \\
224×14×14 & 224×7×7 & - & - & - & - & UpSample \\
224×14×14 & 224×14×14 & LeakyReLu(0.2) & 1 & 1 & 3 & Conv2D \\
224×14×14 & 224×14×14 & LeakyReLu(0.2) & 1 & 1 & 3 & Conv2D \\
224×28×28 & 224×14×14 & - & - & - & - & UpSample \\
56×28×28 & 56×28×28 & LeakyReLu(0.2) & 1 & 1 & 3 & Conv2D \\
56×28×28 & 56×28×28 & LeakyReLu(0.2) & 1 & 1 & 3 & Conv2D \\
56×56×56 & 56×28×28 & - & - & - & - & UpSample \\
28×56×56 & 56×56×56 & LeakyReLu(0.2) & 1 & 1 & 3 & Conv2D \\
28×56×56 & 28×56×56 & LeakyReLu(0.2) & 1 & 1 & 3 & Conv2D \\
28×112×112 & 28×56×56 & - & - & - & - & UpSample \\
14×112×112 & 28×112×112 & LeakyReLu(0.2) & 1 & 1 & 3 & Conv2D \\
14×112×112 & 14×112×112 & LeakyReLu(0.2) & 1 & 1 & 3 & Conv2D \\
14×224×224 & 14×112×112 & - & - & - & - & UpSample \\
7×224×224 & 14×224×224 & LeakyReLu(0.2) & 1 & 1 & 3 & Conv2D \\
7×224×224 & 7×224×224 & LeakyReLu(0.2) & 1 & 1 & 3 & Conv2D \\
3×224×224 & 7×224×224 & Linear & 1 & 0 & 1 & ToRGB \\
\bottomrule
\end{tabular}
\end{table}

The structure of the customized critic network is shown in Table~\ref{tab:critic}:

\begin{table}[H]
\centering
\caption{Structure of the critic network of utilized ProGAN at stage \#6}
\label{tab:critic}
\begin{tabular}{lcccccc}
\toprule
Layer Output Dimensions & Layer Input Dimensions & Activation Functions & S & P & K & Layer Type \\
\midrule
3×224×224 & 3×224×224 & - & - & - & - & Image Input \\
14×224×224 & 3×224×224 & LeakyReLu(0.2) & 1 & 0 & 7 & Conv2D \\
14×224×224 & 14×224×224 & LeakyReLu(0.2) & 1 & 1 & 3 & Conv2D \\
28×112×112 & 14×224×224 & LeakyReLu(0.2) & 1 & 1 & 3 & Conv2D \& DS \\
56×112×112 & 28×112×112 & LeakyReLu(0.2) & 1 & 1 & 3 & Conv2D \\
56×112×112 & 56×112×112 & LeakyReLu(0.2) & 1 & 1 & 3 & Conv2D \\
56×56×56 & 56×112×112 & - & - & - & - & DownSample \\
112×56×56 & 56×56×56 & LeakyReLu(0.2) & 1 & 1 & 3 & Conv2D \\
112×56×56 & 112×56×56 & LeakyReLu(0.2) & 1 & 1 & 3 & Conv2D \\
112×28×28 & 112×56×56 & - & - & - & - & DownSample \\
224×28×28 & 112×28×28 & LeakyReLu(0.2) & 1 & 1 & 3 & Conv2D \\
224×28×28 & 224×28×28 & LeakyReLu(0.2) & 1 & 1 & 3 & Conv2D \\
224×14×14 & 224×28×28 & - & - & - & - & DownSample \\
224×14×14 & 224×14×14 & LeakyReLu(0.2) & 1 & 1 & 3 & Conv2D \\
224×14×14 & 224×14×14 & LeakyReLu(0.2) & 1 & 1 & 3 & Conv2D \\
224×7×7 & 224×14×14 & - & - & - & - & DownSample \\
225×7×7 & 224×7×7 & LeakyReLu(0.2) & - & - & - & MB StdDev \\
224×7×7 & 225×7×7 & LeakyReLu(0.2) & 1 & 1 & 3 & Conv2D \\
224×1×1 & 224×7×7 & LeakyReLu(0.2) & 1 & 1 & 3 & Conv2D \& DS \\
1×1×1 & 224×1×1 & Linear & 1 & 0 & 1 & Output Score \\
\bottomrule
\end{tabular}
\end{table}

According to the structures of the Critic and Generator models, it should be noted that the most inner and outer series of layers (the grayed-out rows) are constant during the progression stages, and only the middle series of layers are added upon finishing the current stage.

\section{RESULTS}

\subsection*{ProGAN Results}

As mentioned in Section 3.4, the utilized ProGAN is trained in 6 stages. Moreover, the only meaningful value which is calculated during the ProGAN training stages are the computed Wasserstein loss. In what follows these values for the critic and generator models are shown (Figure 4 to Figure 9):

\begin{figure}[H]
\centering
\includegraphics[width=0.8\linewidth]{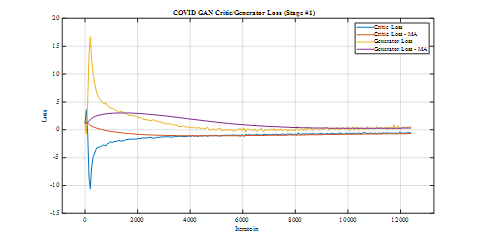}
\caption{Critic \& Generator losses versus iteration at stage \#1 for Covid-19 class}
\label{fig:4}
\end{figure}

According to Figure 4, the computed loss values converged; thus, the capacity and the layers’ structure of the ProGAN are adequate for generating 7×7 images at stage \#1 (it should be noted that merely the convergence of loss values is not sufficient to acclaim the GAN performed well, the quality of the generated images should be considered subsequently).

\begin{figure}[H]
\centering
\includegraphics[width=0.8\linewidth]{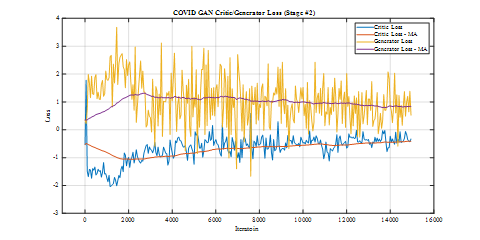}
\caption{Critic \& Generator losses versus iteration at stage \#2 for Covid-19 class}
\label{fig:5}
\end{figure}

In consonance with Figure 5, generator and critic loss values are convergent; therefore, the statement of Figure 4 is valid for the latter figure. The computed Wasserstein loss versus ProGAN training iteration at stage \#3 for covid-19 as label:

\begin{figure}[H]
\centering
\includegraphics[width=0.8\linewidth]{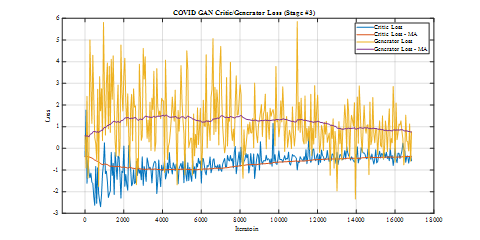}
\caption{Critic \& Generator losses versus iteration at stage \#3 for Covid-19 class}
\label{fig:6}
\end{figure}

The computed Wasserstein loss versus ProGAN training iteration at stage \#5 for covid-19 as label:

\begin{figure}[H]
\centering
\includegraphics[width=0.8\linewidth]{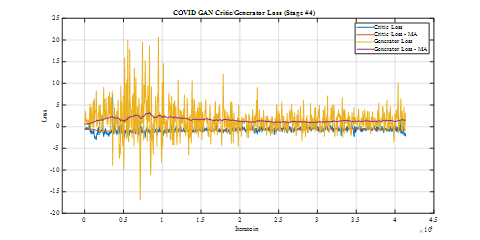}
\caption{Critic \& Generator losses versus iteration at stage \#4 for Covid-19 class}
\label{fig:7}
\end{figure}

The computed Wasserstein loss versus ProGAN training iteration at stage \#5 for covid-19 as label:

\begin{figure}[H]
\centering
\includegraphics[width=0.8\linewidth]{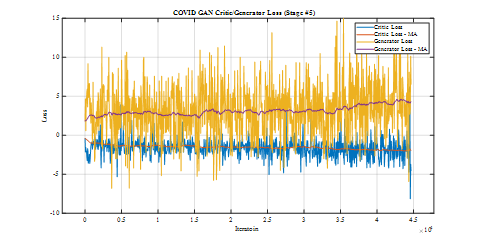}
\caption{Critic \& Generator losses versus iteration at stage \#5 for Covid-19 class}
\label{fig:8}
\end{figure}

The computed Wasserstein loss versus ProGAN training iteration at stage \#6 for covid-19 as label:

\begin{figure}[H]
\centering
\includegraphics[width=0.8\linewidth]{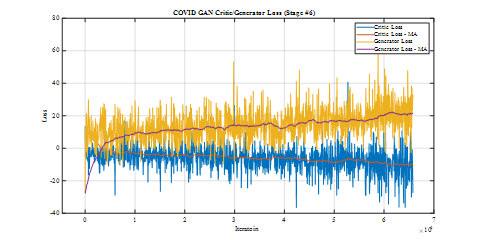}
\caption{Critic \& Generator losses versus iteration at stage \#6 for Covid-19 class}
\label{fig:9}
\end{figure}

The losses versus iterations of other labels are available in the appendix.

According to Figure 4 to Figure 9, in the final stages (Specifically stage \#5 and stage \#6) of training the ProGAN network, the loss values do not tend to be balanced and converge to a specific value. The mentioned point indicates that the networks cannot learn the features of images due to the higher complexity of larger images. The following reasons could be the cause of this issue:
\begin{itemize}
    \item Insufficient capacity of the generator and critic networks
    \item The structure of the generator and critic networks are not expedient
    \item Inappropriate hyper-parameters of the ProGAN network (such as epochs count, batch size, gradient penalty coefficient, and $n_{\text{critic}}$ of each stage of the ProGAN)
\end{itemize}

Despite the not-satisfying results at the final stages, the synthetic images enhanced the performance of ResNet50V2 and VGG16 classifiers. In Figure 10, a sample of synthetic images by the ProGAN with a constant random point from the latent space is shown:

\begin{figure}[H]
\centering
\includegraphics[width=\linewidth]{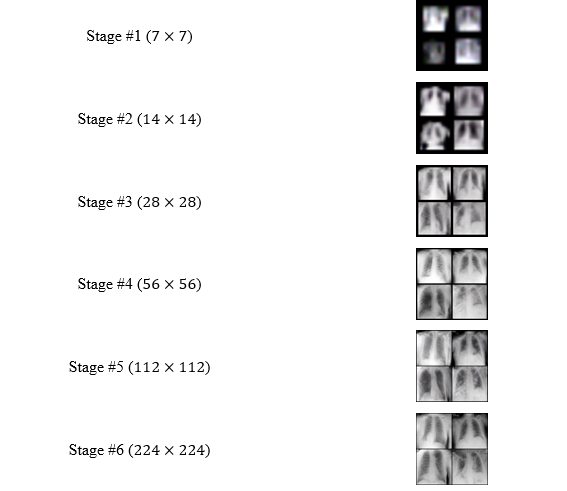}
\caption{Samples of synthetic images generated by ProGAN at a constant random point from latent space (the utilized class is covid-19). Stage \#1 (7×7), Stage \#2 (14×14), Stage \#3 (28×28), Stage \#4 (56×56), Stage \#5 (112×112), Stage \#6 (224×224).}
\label{fig:10}
\end{figure}

Figure 10 shows a sample of synthetic images for the last iteration of each progressive stage. The depicted images correspond to a constant random point in the latent space. According to this figure, the quality of the synthetic images is enriched throughout each progressive stage. At stage \#6, numerous synthetic images are not distinguishable from real images for non-expert individuals.

\subsection*{CLASSIFIER RESULTS}

The optimized hyperparameters of ResNet50V2 found by the SMA optimizer are represented in Table~\ref{tab:sma_opt}.

\begin{table}[H]
\centering
\caption{Optimized value from SMA optimizer}
\label{tab:sma_opt}
\begin{tabular}{lc}
\toprule
Parameters & Optimized Value \\
\midrule
Learning Rate & 7.26e-5 \\
Dropout Rate & 0.17 \\
SIIR & 0.20 \\
\bottomrule
\end{tabular}
\end{table}

In Table~\ref{tab:resnet_exp1}, the outputs of sub-experiments of experiment 1 are represented.

\begin{table}[H]
\centering
\caption{ResNet50V2 outputs – experiments set 1 (reported values are percentages)}
\label{tab:resnet_exp1}
\begin{tabular}{lccccc}
\toprule
(Class/Dataset/SIIR) & Recall$\pm$STD & Specificity$\pm$STD & F1Score$\pm$STD & Precision$\pm$STD & Accuracy$\pm$STD \\
\midrule
CNVL/FullDS/0.00 & 92.00$\pm$0.28 & 97.33$\pm$0.09 & 91.97$\pm$0.28 & 92.02$\pm$0.28 & 92.00$\pm$0.28 \\
CNVL/FullDS/0.20 (*) & 95.53$\pm$1.57 & 95.51$\pm$0.08 & 95.49$\pm$0.25 & 95.57$\pm$0.25 & 95.53$\pm$1.57 \\
CN/FullDS/0.00 & 97.00$\pm$0.17 & 97.00$\pm$0.17 & 96.98$\pm$0.17 & 96.99$\pm$0.17 & 97.00$\pm$0.17 \\
CN/FullDS/0.20 (*) & 98.49$\pm$0.14 & 98.49$\pm$0.14 & 98.49$\pm$0.15 & 98.48$\pm$0.14 & 98.50$\pm$0.14 \\
CN/FullDS/0.15 & 97.41$\pm$0.15 & 97.41$\pm$0.15 & 97.40$\pm$0.15 & 97.40$\pm$0.15 & 97.41$\pm$0.15 \\
CN/FullDS/0.10 & 97.44$\pm$0.13 & 97.44$\pm$0.13 & 97.43$\pm$0.13 & 97.43$\pm$0.13 & 97.44$\pm$0.13 \\
CN/FullDS/0.05 & 97.21$\pm$0.25 & 97.21$\pm$0.25 & 97.19$\pm$0.26 & 97.20$\pm$0.25 & 97.21$\pm$0.25 \\
CN/2048/0.00 & 90.89$\pm$0.63 & 90.89$\pm$0.63 & 90.89$\pm$0.62 & 90.99$\pm$0.63 & 90.89$\pm$0.63 \\
CN/2048/0.20 & 98.21$\pm$0.77 & 98.21$\pm$0.77 & 98.21$\pm$0.77 & 98.30$\pm$0.77 & 98.21$\pm$0.77 \\
CN/2048/0.15 & 94.32$\pm$0.56 & 94.32$\pm$0.56 & 94.31$\pm$0.55 & 94.37$\pm$0.57 & 94.32$\pm$0.56 \\
CN/2048/0.10 & 91.07$\pm$1.00 & 91.07$\pm$1.00 & 91.07$\pm$1.00 & 91.19$\pm$1.02 & 91.07$\pm$1.00 \\
CN/2048/0.05 & 92.07$\pm$0.83 & 92.07$\pm$0.83 & 92.07$\pm$0.84 & 92.20$\pm$0.81 & 92.07$\pm$0.83 \\
CN/2048/0.4 & 99.20$\pm$0.19 & 99.20$\pm$0.19 & 99.20$\pm$0.19 & 99.20$\pm$0.19 & 99.20$\pm$0.19 \\
\bottomrule
\end{tabular}

\vspace{0.5em}
\small (*): The optimized values are used to train the ResNet50V2
\end{table}

\begin{table}[H]
\centering
\caption{VGG16 outputs – experiments set 2 (reported values are percentages)}
\label{tab:vgg_exp2}
\begin{tabular}{lccccc}
\toprule
(Classes/Dataset/SIIR) & Recall$\pm$STD & Specificity$\pm$STD & F1Score$\pm$STD & Precision$\pm$STD & Accuracy$\pm$STD \\
\midrule
CNVL/FullDS/0.90 & 89.97$\pm$1.97 & 96.68$\pm$0.74 & 89.23$\pm$2.63 & 89.03$\pm$2.19 & 88.13$\pm$2.20 \\
CNVL/FullDS/0.80 & 89.95$\pm$2.27 & 96.55$\pm$0.81 & 89.95$\pm$2.56 & 88.25$\pm$2.46 & 88.45$\pm$2.26 \\
CNVL/FullDS/0.70 & 89.55$\pm$2.07 & 96.52$\pm$0.69 & 89.45$\pm$2.07 & 89.15$\pm$2.07 & 88.55$\pm$2.07 \\
CNVL/FullDS/0.60 & 87.87$\pm$2.87 & 95.94$\pm$0.96 & 87.82$\pm$2.87 & 87.42$\pm$2.87 & 87.82$\pm$2.87 \\
CNVL/FullDS/0.50 & 88.95$\pm$1.31 & 96.15$\pm$0.44 & 88.85$\pm$1.31 & 88.25$\pm$1.31 & 88.45$\pm$1.31 \\
CNVL/FullDS/0.40 & 89.81$\pm$1.44 & 96.34$\pm$0.48 & 89.51$\pm$1.44 & 89.01$\pm$1.44 & 89.01$\pm$1.44 \\
CNVL/FullDS/0.30 & 89.96$\pm$1.45 & 96.39$\pm$0.48 & 89.46$\pm$1.45 & 89.66$\pm$1.45 & 89.16$\pm$1.45 \\
CNVL/FullDS/0.20 & 89.98$\pm$2.42 & 96.43$\pm$0.81 & 89.56$\pm$2.42 & 89.28$\pm$2.42 & 89.28$\pm$2.42 \\
CNVL/FullDS/0.10 & 88.69$\pm$2.30 & 96.22$\pm$0.77 & 88.65$\pm$2.30 & 88.15$\pm$2.31 & 90.14$\pm$2.25 \\
CNVL/FullDS/0.00 & 89.90$\pm$2.08 & 96.50$\pm$0.69 & 89.87$\pm$2.08 & 89.52$\pm$2.08 & 89.50$\pm$2.08 \\
\bottomrule
\end{tabular}
\end{table}

\begin{figure}[H]
\centering
\includegraphics[width=0.95\linewidth]{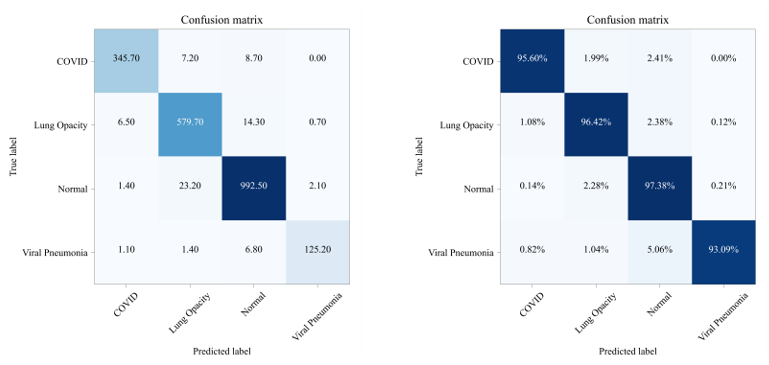}
\caption{Average confusion matrix computed from cross-validation (left side). The normalized averaged confusion matrix (right side), Problem: 4-class classification of the full imbalanced real dataset using the optimized ResNet50V2}
\label{fig:11}
\end{figure}

Figure 11 represents the normalized and averaged computed confusion matrices from cross-validation. According to this figure, the classification accuracy of the Viral Pneumonia class is at least one, and the Normal class has the highest accuracy compared to other classes. This fact is expected since the frequency of the latter, and the former classes are the lowest and the highest, respectively. 

\begin{figure}[H]
\centering
\includegraphics[width=0.95\linewidth]{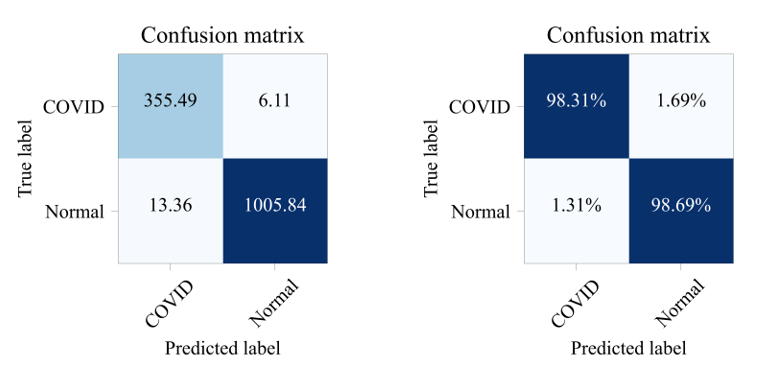}
\caption{Average confusion matrix computed from cross-validation (left side). The normalized averaged confusion matrix (right side), Problem: 2-class classification of the full imbalanced real dataset}
\label{fig:12}
\end{figure}

Furthermore, in Figure 12, the normalized and averaged computed confusion matrices from cross-validation are represented. It should be noted that the classifier for this problem is not optimized; however, the same optimized values found for hitherto 4-class classification are used in this problem as well.

\begin{figure}[H]
\centering
\includegraphics[width=0.95\linewidth]{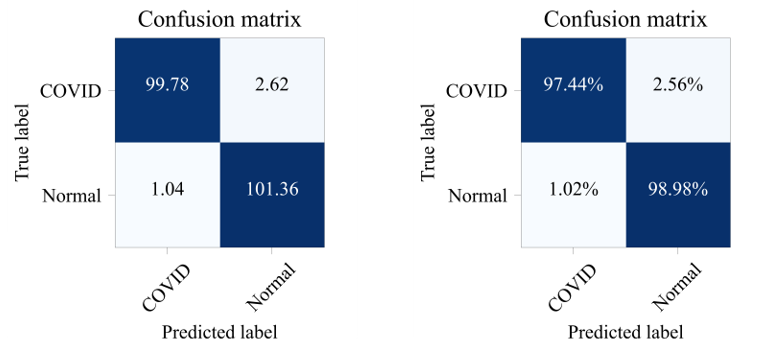}
\caption{Average confusion matrix computed from cross-validation (left side). The normalized averaged confusion matrix (right side), Problem: 2-class classification of 2048 randomly selected images for training the ProGANs}
\label{fig:13}
\end{figure}

Finally, in Figure 13, the normalized and averaged computed confusion matrices from cross-validation are represented. The investigated problem in this scenario is classifying 2048 randomly selected images used for training ProGANs (1024 images for each label).

\section{CONCLUSION}

In this study, an effort is made to answer the challenge regarding the scarcity of data and imbalance of the most extensive chest-X ray available dataset. Therefore, four ProGAN networks are distinctively trained for each class: COVID, Viral Pneumonia, Lung Opacity, and Normal. As the next step, a synthetic images dataset containing 40K CXR images is generated using the trained ProGAN networks. Successively, by utilizing the synthetic dataset and concatenating it with the real dataset, the imbalance of the real dataset is lessened. Throughout this procedure, an SMA algorithm is used to optimize the SIIR, Learning Rate, and Drop Out rate of the pre-trained ResNet50V2 classifier network. A ResNet50V2 is fully trained to address the classification problem using the conclusive optimized hyper-parameters. To further scrutinize the impact of adding the synthetic images and using the optimizer, several experiments were conducted to compare the cross-validated computed metrics of various synthetic image injection rates for classifiers (the experiments are based on ResNet50V2 and VGG16 networks). It should be noted that a weighted approach is taken into account for adding the synthetic images to the real dataset; as a result, the lower the frequency of the images of a specific class (which indicates a higher imbalance ratio), the higher number of synthetic images are added to the real dataset with the same class.

In essence, the following results are obtained:
\begin{itemize}
    \item The best injection rate of synthetic images to the real dataset is 20\%.
    \item The best learning rate for training the ResNet50V2 network is 7.26e-5
    \item The generated synthetic images acquired from the four trained ProGAN networks alongside the optimized hyper-parameters enhanced classification accuracy by 3.53\% for the ResNet50V2 network.
    \item The reduction of standard deviation for the experiment with synthetic images and optimized hyper-parameters is an indication of reduction in the robustness of the classification task
    \item The contemporary ProGAN networks are able to generate useful synthetic images for enhancing the models’ classification metrics
\end{itemize}

\bibliographystyle{unsrt}

\begin{thebibliography}{150}

\bibitem{WHO_SARS}
Severe Acute Respiratory Syndrome (SARS). \url{https://www.who.int/health-topics/severe-acute-respiratory-syndrome#tab=tab_1} (accessed Jul. 05, 2022).

\bibitem{WHO_COVID_timeline}
Timeline of WHO’s response to COVID-19. \url{https://www.who.int/emergencies/diseases/novel-coronavirus-2019/interactive-timeline#event-0} (accessed Apr. 15, 2022).

\bibitem{CDC_COVID_timeline}
CDC Museum COVID-19 Timeline | David J. Sencer CDC Museum | CDC. \url{https://www.cdc.gov/museum/timeline/covid19.html} (accessed Jul. 06, 2022).

\bibitem{Worldometers_COVID}
Worldometers, COVID-19 CORONAVIRUS PANDEMIC. \url{https://www.worldometers.info/coronavirus/}.

\bibitem{Worldometers_graphs}
Coronavirus Graphs: Worldwide Cases and Deaths - Worldometer. \url{https://www.worldometers.info/coronavirus/worldwide-graphs/#countries-cases} (accessed Apr. 15, 2022).

\bibitem{GomezCarballa2021}
A. Gómez-Carballa, J. Pardo-Seco, X. Bello, F. Martinón-Torres, and A. Salas, Superspreading in the emergence of COVID-19 variants, Trends Genet., vol. 37, no. 12, pp. 1069–1080, Dec. 2021, doi: 10.1016/J.TIG.2021.09.003.

\bibitem{Araf2022}
Y. Araf et al., Omicron variant of SARS-CoV-2: Genomics, transmissibility, and responses to current COVID-19 vaccines, J. Med. Virol., vol. 94, no. 5, pp. 1825–1832, May 2022, doi: 10.1002/JMV.27588.

\bibitem{Bouzid2022}
D. Bouzid et al., Comparison of Patients Infected With Delta Versus Omicron COVID-19 Variants Presenting to Paris Emergency Departments, https://doi.org/10.7326/M22-0308, vol. 175, no. 6, pp. 831–837, Mar. 2022, doi: 10.7326/M22-0308.

\bibitem{Iacobucci2021}
G. Iacobucci, Covid-19: Runny nose, headache, and fatigue are commonest symptoms of omicron, early data show, BMJ, vol. 375, p. n3103, Dec. 2021, doi: 10.1136/BMJ.N3103.

\bibitem{Mahase2021}
E. Mahase, Covid-19: Sore throat, fatigue, and myalgia are more common with new UK variant, BMJ, vol. 372, p. n288, Jan. 2021, doi: 10.1136/BMJ.N288.

\bibitem{WHO_coronavirus}
Coronavirus. \url{https://www.who.int/health-topics/coronavirus#tab=tab_3} (accessed Jul. 06, 2022).

\bibitem{Arjovsky2017WGAN}
M. Arjovsky, S. Chintala, and L. Bottou, Wasserstein GaN, arXiv, 2017.

\bibitem{Mahdizadehaghdam2019SparseGAN}
S. Mahdizadehaghdam, A. Panahi, and H. Krim, Sparse generative adversarial network, Proc. - 2019 Int. Conf. Comput. Vis. Work. ICCVW 2019, pp. 3063–3071, 2019, doi: 10.1109/ICCVW.2019.00369.

\bibitem{Gulrajani2017ImprovedWGAN}
I. Gulrajani, Improved Training of Wasserstein GANs.

\bibitem{Karras2018ProGAN}
J. L. Tero Karras, Timo Aila, Samuli Laine, Progressive Growing of GANs for Improved Quality, Stability, and Variation, ICLR 2018, 2018. \url{https://research.nvidia.com/publication/2018-04_progressive-growing-gans-improved-quality-stability-and-variation} (accessed May 30, 2022).

\bibitem{Roy2022SVDCLAHE}
S. Roy, M. Tyagi, V. Bansal, and V. Jain, SVD-CLAHE boosting and balanced loss function for Covid-19 detection from an imbalanced Chest X-Ray dataset, Comput. Biol. Med., vol. 150, p. 106092, Nov. 2022, doi: 10.1016/J.COMPBIOMED.2022.106092.

\bibitem{Calderon2021MixMatch}
S. Calderon-Ramirez et al., Correcting data imbalance for semi-supervised COVID-19 detection using X-ray chest images, Appl. Soft Comput., vol. 111, p. 107692, Nov. 2021, doi: 10.1016/J.ASOC.2021.107692.

\bibitem{Nour2020DeepFeatures}
M. Nour, Z. Cömert, and K. Polat, A Novel Medical Diagnosis model for COVID-19 infection detection based on Deep Features and Bayesian Optimization, Appl. Soft Comput., vol. 97, no. xxxx, p. 106580, 2020, doi: 10.1016/j.asoc.2020.106580.

\bibitem{Wan2018DR}
S. Wan, Y. Liang, and Y. Zhang, Deep convolutional neural networks for diabetic retinopathy detection by image classification, Comput. Electr. Eng., vol. 72, pp. 274–282, 2018, doi: 10.1016/j.compeleceng.2018.07.042.

\bibitem{Zunair2021CycleGAN}
H. Zunair and A. Ben Hamza, Synthesis of COVID-19 chest X-rays using unpaired image-to-image translation, Soc. Netw. Anal. Min., vol. 11, no. 1, pp. 1–12, Dec. 2021, doi: 10.1007/S13278-021-00731-5/FIGURES/8.

\bibitem{Panahi2021FCOD}
A. H. Panahi, A. Rafiei, and A. Rezaee, FCOD: Fast COVID-19 Detector based on deep learning techniques, Informatics Med. Unlocked, vol. 22, p. 100506, 2021, doi: 10.1016/j.imu.2020.100506.

\bibitem{Zhao2021EHSSA}
S. Zhao, P. Wang, A. A. Heidari, H. Chen, W. He, and S. Xu, Performance optimization of salp swarm algorithm for multi-threshold image segmentation: Comprehensive study of breast cancer microscopy, Comput. Biol. Med., vol. 139, Dec. 2021, doi: 10.1016/j.compbiomed.2021.105015.

\bibitem{Rahman2021Enhancement}
T. Rahman et al., Exploring the effect of image enhancement techniques on COVID-19 detection using chest X-ray images, Comput. Biol. Med., vol. 132, no. March, p. 104319, 2021, doi: 10.1016/j.compbiomed.2021.104319.

\bibitem{Haghanifar2022COVIDCXNet}
A. Haghanifar, M. M. Majdabadi, and S. Ko, COVID-CXNet: Detecting covid-19 in frontal chest x-ray images using deep learning, arXiv, 2020.

\bibitem{Badawi2021Balanced}
A. Badawi and K. Elgazzar, Detecting Coronavirus from Chest X-rays Using Transfer Learning, COVID 2021, Vol. 1, Pages 403-415, vol. 1, no. 1, pp. 403–415, Sep. 2021, doi: 10.3390/COVID1010034.

\bibitem{Wang2019ChestXray}
X. Wang, Y. Peng, L. Lu, Z. Lu, M. Bagheri, and R. M. Summers, ChestX-ray: Hospital-Scale Chest X-ray Database and Benchmarks on Weakly Supervised Classification and Localization of Common Thorax Diseases, in Advances in Computer Vision and Pattern Recognition, Jul. 2019, pp. 369–392. doi: 10.1007/978-3-030-13969-8\_18.

\bibitem{Kermany2018Cell}
D. S. Kermany et al., Identifying Medical Diagnoses and Treatable Diseases by Image-Based Deep Learning, Cell, vol. 172, no. 5, pp. 1122-1131.e9, 2018, doi: 10.1016/j.cell.2018.02.010.

\bibitem{Winther2020ImageRepo}
B. C. Winther, Hinrich B. and Laser, Hans and Gerbel, Svetlana and Maschke, Sabine K. and B. Hinrichs, Jan and Vogel-Claussen, Jens and Wacker, Frank K. and Höper, Marius M. and Meyer, COVID-19 Image Repository, Figshare, p. May, 2020, [Online]. Available: \url{https://figshare.com/articles/dataset/COVID-19_Image_Repository/12275009/1}.

\bibitem{Vaya2020BIMCV}
M. de la I. Vayá et al., BIMCV COVID-19+: a large annotated dataset of RX and CT images from COVID-19 patients, arXiv Prepr. arXiv2006.01174, 2020, [Online]. Available: \url{http://arxiv.org/abs/2006.01174}.

\bibitem{Cohen2020Dataset}
J. P. Cohen, P. Morrison, and L. Dao, COVID-19 Image Data Collection. 2020. [Online]. Available: \url{http://arxiv.org/abs/2003.11597}.

\bibitem{Shams2020Mendeley}
M. Shams, O. Elzeki, M. Abd Elfattah, and A. Hassanien, Chest x-ray images with three classes: covid-19, normal, and pneumonia, Mendeley Data v3. 2020.

\bibitem{Chung2020Figure1}
A. Chung, Figure1-COVID-chestxray-dataset, GitHub repository. GitHub, 2020.

\bibitem{Chung2020ActualMed}
A. Chung, Actualmed-COVID-chestxray-dataset, GitHub repository. GitHub, 2020.

\bibitem{ItalianSocietyRadiology2021}
COVID-19, Italian Society of Medical and Interventional Radiology. Italian Society of Medical and Interventional Radiology, 2021.

\bibitem{TwitterChestImaging}
Chest Imaging. Twitter, 2021.

\bibitem{Goodfellow2014GAN}
I. Goodfellow et al., Generative Adversarial Nets, in Advances in Neural Information Processing Systems, 2014, vol. 27. [Online]. Available: \url{https://proceedings.neurips.cc/paper/2014/file/5ca3e9b122f61f8f06494c97b1afccf3-Paper.pdf}.

\bibitem{Odena2017ACGAN}
A. Odena, C. Olah, J. Shlens, and G. Brain, CONDITIONAL IMAGE SYNTHESIS WITH AUXILIARY CLASSIFIER GANS, pp. 1–16, 2017.

\bibitem{Chen2016InfoGAN}
X. Chen, Y. Duan, R. Houthooft, J. Schulman, I. Sutskever, and P. Abbeel, InfoGAN: Interpretable Representation Learning by Information Maximizing Generative Adversarial Nets.

\bibitem{Mirza2014CGAN}
M. Mirza and S. Osindero, Conditional Generative Adversarial Nets, Nov. 2014, doi: 10.48550/arxiv.1411.1784.

\bibitem{Chowdhury2020AI}
M. E. H. Chowdhury et al., Can AI help in screening Viral and COVID-19 pneumonia?, 2020, doi: 10.1109/ACCESS.2020.3010287.

\bibitem{WongYeh2020Kfold}
T. T. Wong and P. Y. Yeh, Reliable Accuracy Estimates from k-Fold Cross Validation, IEEE Trans. Knowl. Data Eng., vol. 32, no. 8, pp. 1586–1594, Aug. 2020, doi: 10.1109/TKDE.2019.2912815.

\bibitem{Pathan2021}
S. Pathan, P. C. Siddalingaswamy, and T. Ali, Automated Detection of Covid-19 from Chest X-ray scans using an optimized CNN architecture, Appl. Soft Comput., vol. 104, p. 107238, 2021, doi: 10.1016/j.asoc.2021.107238.

\bibitem{Panwar2020nCOVnet}
H. Panwar, P. K. Gupta, M. K. Siddiqui, R. Morales-Menendez, P. Bhardwaj, and V. Singh, A deep learning and grad-CAM based color visualization approach for fast detection of COVID-19 cases using chest X-ray and CT-Scan images, Chaos, Solitons and Fractals, vol. 140, p. 110190, 2020, doi: 10.1016/j.chaos.2020.110190.

\bibitem{Goel2021OptCoNet}
T. Goel, R. Murugan, S. Mirjalili, and D. K. Chakrabartty, OptCoNet: an optimized convolutional neural network for an automatic diagnosis of COVID-19, Appl. Intell. (Dordrecht, Netherlands), vol. 51, no. 3, pp. 1351–1366, Mar. 2021, doi: 10.1007/S10489-020-01904-Z.

\bibitem{Ozturk2020DarkCovidNet}
T. Ozturk, M. Talo, E. A. Yildirim, U. B. Baloglu, O. Yildirim, and U. Rajendra Acharya, Automated detection of COVID-19 cases using deep neural networks with X-ray images, Comput. Biol. Med., vol. 121, p. 103792, Jun. 2020, doi: 10.1016/J.COMPBIOMED.2020.103792.

\bibitem{Hussain2021CoroDet}
E. Hussain, M. Hasan, M. A. Rahman, I. Lee, T. Tamanna, and M. Z. Parvez, CoroDet: A deep learning based classification for COVID-19 detection using chest X-ray images, Chaos, Solitons and Fractals, vol. 142, p. 110495, 2021, doi: 10.1016/j.chaos.2020.110495.

\bibitem{Waheed2021CovidGAN}
A. Waheed, M. Goyal, D. Gupta, A. Khanna, F. Al-Turjman, and P. R. Pinheiro, CovidGAN: Data Augmentation Using Auxiliary Classifier GAN for Improved Covid-19 Detection, IEEE Access, vol. 8, pp. 91916–91923, Mar. 2021, doi: 10.1109/ACCESS.2020.2994762.

\bibitem{Ahamed2021DeepPreprocess}
K. U. Ahamed et al., A deep learning approach using effective preprocessing techniques to detect COVID-19 from chest CT-scan and X-ray images, Comput. Biol. Med., vol. 139, p. 105014, Dec. 2021, doi: 10.1016/J.COMPBIOMED.2021.105014.


\bibitem{AbdoliHajati2014}
S. Abdoli and F. Hajati, Offline signature verification using geodesic derivative pattern, in 2014 22nd Iranian Conference on Electrical Engineering (ICEE), 2014: IEEE, pp. 1018--1023.

\bibitem{Ayatollahi2015}
F. Ayatollahi, A. A. Raie, and F. Hajati, Expression-invariant face recognition using depth and intensity dual-tree complex wavelet transform features, Journal of Electronic Imaging, vol. 24, no. 2, pp. 023031-023031, 2015.

\bibitem{BarolliAINA2024}
L. Barolli, Advanced Information Networking and Applications: Proceedings of the 38th International Conference on Advanced Information Networking and Applications (AINA-2024), Volume 2, ed: Springer Nature, 2024.

\bibitem{BarolliBWCCA2019}
L. Barolli, P. Hellinckx, and T. Enokido, Advances on Broad-Band Wireless Computing, Communication and Applications: Proceedings of the 14th International Conference on Broad-Band Wireless Computing, Communication and Applications (BWCCA-2019). Springer Nature, 2019.

\bibitem{BarolliWAINA2019}
L. Barolli, M. Takizawa, F. Xhafa, and T. Enokido, Web, Artificial Intelligence and Network Applications: Proceedings of the Workshops of the 33rd International Conference on Advanced Information Networking and Applications (WAINA-2019). Springer, 2019.

\bibitem{Barzamini2012}
R. Barzamini, F. Hajati, S. Gheisari, and M. Motamadinejad, Short term load forecasting using multi-layer perception and fuzzy inference systems, Journal of Applied Sciences, vol. 12, no. 1, pp. 40-47, 2012.

\bibitem{CremersACCV2014}
D. Cremers, I. Reid, H. Saito, and M.-H. Yang, Computer Vision--ACCV 2014: 12th Asian Conference on Computer Vision, Singapore, Singapore, November 1-5, 2014, Revised Selected Papers, Part V. Springer, 2015.

\bibitem{Fiorini2019}
S. Fiorini, F. Hajati, A. Barla, and F. Girosi, Predicting diabetes second-line therapy initiation in the Australian population via timespan-guided neural attention network, PLOS ONE, vol. 10, no. 14, p. e0211844, 2019.

\bibitem{Hajati2017Surface}
F. Hajati, A. Cheraghian, S. Gheisari, Y. Gao, and A. S. Mian, Surface geodesic pattern for 3D deformable texture matching, Pattern Recognition, vol. 62, pp. 21-32, 2017.

\bibitem{Hajati2006FaceLocalization}
F. Hajati, K. Faez, and S. K. Pakazad, An Efficient Method for Face Localization and Recognition in Color Images, in Systems, Man and Cybernetics, 2006. SMC'06. IEEE International Conference on, 2006, vol. 5: IEEE, pp. 4214-4219.

\bibitem{Hajati2010PoseInvariant}
F. Hajati, A. A. Raie, and Y. Gao, Pose-invariant 2.5 D face recognition using geodesic texture warping, in 2010 11th International Conference on Control Automation Robotics \& Vision, 2010: IEEE, pp. 1837-1841.

\bibitem{Hajati2017DynamicTexture}
F. Hajati, M. Tavakolian, S. Gheisari, Y. Gao, and A. S. Mian, Dynamic texture comparison using derivative sparse representation: Application to video-based face recognition, IEEE Transactions on Human-Machine Systems, vol. 47, no. 6, pp. 970-982, 2017.

\bibitem{Mahajan2024_ens}
P. Mahajan, S. Uddin, F. Hajati, M. A. Moni, and E. Gide, A comparative evaluation of machine learning ensemble approaches for disease prediction using multiple datasets, Health and Technology, vol. 14, no. 3, pp. 597-613, 2024.

\bibitem{Pakazad2006FaceDetection}
S. K. Pakazad, K. Faez, and F. Hajati, Face Detection Based on Central Geometrical Moments of Face Components, in Systems, Man and Cybernetics, 2006. SMC'06. IEEE International Conference on, 2006, pp. 4225-4230.

\bibitem{Sadeghi2024COVID_new}
A. Sadeghi, M. Sadeghi, A. Sharifpour, M. Fakhar, Z. Zakariaei, and F. Hajati, Potential diagnostic application of a novel deep learning-based approach for COVID-19, Nature Scientific Reports, vol. 14, no. 1, p. 280, 2024.

\bibitem{Shojaiee2014Palmprint}
F. Shojaiee and F. Hajati, Local composition derivative pattern for palmprint recognition, in 2014 22nd Iranian Conference on Electrical Engineering (ICEE), 2014: IEEE, pp. 965-970.

\bibitem{Sopo2021DeFungi}
C. J. P. Sopo, F. Hajati, and S. Gheisari, DeFungi: Direct mycological examination of microscopic fungi images, arXiv preprint arXiv:2109.07322, 2021.

\bibitem{Tavakolian2022FastCOVID_new}
A. Tavakolian, F. Hajati, A. Rezaee, A. O. Fasakhodi, and S. Uddin, Fast COVID-19 versus H1N1 screening using optimized parallel inception, Expert systems with applications, vol. 204, p. 117551, 2022.

\bibitem{Tavakolian2023Readmission_new}
A. Tavakolian, A. Rezaee, F. Hajati, and S. Uddin, Hospital readmission and length-of-stay prediction using an optimized hybrid deep model, Future Internet, vol. 15, no. 9, p. 304, 2023.

\bibitem{Wang2022SoftwareImpacts_new}
S. Wang, H. Lu, A. Khan, F. Hajati, M. Khushi, and S. Uddin, A machine learning software tool for multiclass classification, Software Impacts, vol. 13, p. 100383, 2022.

\bibitem{KarimiRezaee2017Helmholtz}
M. Karimi and A. Rezaee, Regularization of the Cauchy problem for the Helmholtz equation by using Meyer wavelet, Journal of Computational and Applied Mathematics, vol. 320, pp. 76-95, 2017.

\bibitem{MohamadzadeRezaee2017Antenna}
B. Mohamadzade and A. Rezaee, Compact and broadband dual sleeve monopole antenna for GSM, WiMAX and WLAN application, Microwave and Optical Technology Letters, vol. 59, no. 6, pp. 1271-1277, 2017.

\bibitem{Ramezani2024Drones}
M. Ramezani, M. Amiri Atashgah, and A. Rezaee, A Fault-Tolerant Multi-Agent Reinforcement Learning Framework for Unmanned Aerial Vehicles–Unmanned Ground Vehicle Coverage Path Planning, Drones, vol. 8, no. 10, p. 537, 2024.

\bibitem{Rezaee2008GeneticSymbiosis}
A. Rezaee, Genetic symbiosis algorithm generating test data for constraint automata, Applied and Computational Mathematics, vol. 6, no. 1, pp. 126-137, 2008.

\bibitem{Rezaee2010FIR}
A. Rezaee, Using Genetic Algorithms for Designing of FIR Digital Filters, ICTACT journal on Soft computing, vol. 1, no. 1, pp. 18-22, 2010.

\bibitem{Rezaee2017PID}
A. Rezaee, Determining PID controller coefficients for the moving motor of a welder robot using fuzzy logic, Automatic Control and Computer Sciences, vol. 51, no. 2, pp. 124-132, 2017.

\bibitem{Rezaee2017Penetrometer}
A. Rezaee, DESIGN, CONSTRUCTION AND EVALUATION OF A DIGITAL HAND-PUSHED PENETROMETER, International Journal of Advanced Smart Sensor Network Systems, vol. 7, no. 1, pp. 1-10, 2017.

\bibitem{Rezaee2017MPC}
A. Rezaee, Model predictive for Mobile robot control, Transactions on environment and electrical engineering, vol. 2, no. 2, pp. 17-22, 2017.

\bibitem{RezaeeGolpayegani2012}
A. Rezaee and M. K. Golpayegani, Intelligent Control of Cooling-Heating Systems by Using Emotional Learning, Electronics and electrical engineering, vol. 18, no. 4, pp. 26-30, 2012.

\bibitem{RezaeePajohesh2016}
A. Rezaee and M. Pajohesh, Suspension system control with fuzzy logic, Journal of Communications Technology, Electronics and Computer Science, vol. 6, pp. 1-5, 2016.

\bibitem{Sadeghi2024ECG}
A. Sadeghi, F. Hajati, A. Rezaee, M. Sadeghi, A. Argha, and H. Alinejad-Rokny, 3DECG-Net: ECG fusion network for multi-label cardiac arrhythmia detection, Computers in Biology and Medicine, vol. 182, p. 109126, 2024.

\bibitem{Taghvaee2014Metamaterial}
H. Taghvaee, S. M. Seyyedi, and A. Rezaee, Design of Metamaterial Dual Band Absorber, in The third Iranian Conference on Engineering Electromagnetic, 2014.

\bibitem{Tavakolian2022SoftwareImpacts_new}
A. Tavakolian, F. Hajati, A. Rezaee, A. O. Fasakhodi, and S. Uddin, Source code for optimized parallel inception: A fast COVID-19 screening software, Software Impacts, vol. 13, p. 100337, 2022.

\bibitem{Gavagsaz2018LoadBalancing}
E. Gavagsaz, A. Rezaee, and H. Haj Seyyed Javadi, Load balancing in reducers for skewed data in MapReduce systems by using scalable simple random sampling, The Journal of Supercomputing, vol. 74, no. 7, pp. 3415-3440, 2018.

\bibitem{Rezaee2014FuzzyCloud}
A. Rezaee, A. M. Rahmani, A. Movaghar, and M. Teshnehlab, Formal process algebraic modeling, verification, and analysis of an abstract Fuzzy Inference Cloud Service, The Journal of Supercomputing, vol. 67, no. 2, pp. 345-383, 2014.

\bibitem{Sarvghad2011ThinkingStyles}
S. Sarvghad, A. Rezaee, and F. Masomi, On the Relationship between Thinking Styles and Self-Efficacy of Pre-University Students in Shiraz, 2011.

\bibitem{Shahramian2013Leptin}
I. Shahramian, E. Akhlaghi, A. Ramezani, A. Rezaee, N. Noori, and E. Sharafi, A study of leptin serum concentrations in patients with major beta-thalassemia, Iranian journal of pediatric hematology and oncology, vol. 3, no. 2, p. 59, 2013.

\bibitem{Shahramian2013Troponin}
I. Shahramian, M. Razzaghian, A. A. Ramazani, G. A. Ahmadi, N. M. Noori, and A. R. Rezaee, The correlation between troponin and ferritin serum levels in the patients with major beta-thalassemia, International cardiovascular research journal, vol. 7, no. 2, p. 51, 2013.

\end{thebibliography}

\end{document}